\documentclass[sigconf,nonacm]{acmart}

\usepackage{amsmath}
\usepackage{booktabs}
\usepackage{graphicx}
\graphicspath{{figures/}}
\usepackage{microtype}
\usepackage{multirow}
\usepackage{tikz}
\usetikzlibrary{arrows.meta,positioning,shapes.geometric,
                fit,backgrounds,calc,decorations.pathreplacing}
\usepackage{enumitem}

\setcopyright{none}
\acmConference[]{}{}{}
\acmBooktitle{}
\acmDOI{}
\acmISBN{}

\begin{document}

\title{Neuromorphic Continual Learning for Sequential Deployment\\
  of Nuclear Plant Monitoring Systems}

\author{Samrendra Roy}
\email{roysam@illinois.edu}
\affiliation{%
  \institution{University of Illinois Urbana-Champaign}
  \department{Department of Nuclear, Plasma \& Radiological Engineering}
  \city{Urbana}
  \state{IL}
  \country{USA}
}

\author{Sajedul Talukder}
\affiliation{%
  \institution{University of Texas at El Paso}
  \department{Department of Computer Science}
  \city{El Paso}
  \state{TX}
  \country{USA}
}

\author{Syed Bahauddin Alam}
\affiliation{%
  \institution{University of Illinois Urbana-Champaign}
  \department{Department of Nuclear, Plasma \& Radiological Engineering}
  \city{Urbana}
  \state{IL}
  \country{USA}
}
\affiliation{%
  \institution{National Center for Supercomputing Applications}
  \city{Urbana}
  \state{IL}
  \country{USA}
}


\begin{abstract}
Anomaly detection in nuclear industrial control systems (ICS)
requires continuous, energy-efficient monitoring across multiple
subsystems that are often deployed at different stages of plant
commissioning.  When a conventional neural network is sequentially
trained to monitor new subsystems, it catastrophically forgets
previously learned anomaly patterns, a safety-critical failure
mode.  We present the first spiking neural network (SNN)-based
anomaly detection system with continual learning for nuclear ICS,
addressing both challenges simultaneously.  Our approach
introduces \emph{spike-encoded asynchronous sensor fusion}, a
delta-based encoding that converts heterogeneous sensor streams into
sparse spike trains at rates dictated by each sensor's natural
dynamics, achieving 92.7\% input sparsity.  We evaluate five continual
learning strategies, including sequential fine-tuning, Elastic Weight
Consolidation (EWC), Synaptic Intelligence (SI), experience replay,
and a hybrid EWC+Replay approach, on the HAI~21.03 nuclear ICS
security dataset across three sequentially deployed subsystems
(boiler, turbine, water treatment).  The hybrid EWC+Replay method
achieves an average F1 score of 0.979 with near-zero average forgetting
(AF\,=\,0.000 single seed; $0.035 \pm 0.039$ across three seeds),
while requiring 12.6$\times$ fewer operations (an estimated
2.5$\times$ in energy based on published hardware specifications)
than an equivalent artificial neural network.  The system detects all tested
attacks with a mean latency of 0.6 seconds.  These results demonstrate
that neuromorphic computing offers a viable path toward
always-on, energy-efficient, and adaptable safety monitoring for
next-generation nuclear facilities.
\end{abstract}


\maketitle

\noindent\textit{Preprint --- submitted to ACM International Conference on
Neuromorphic Systems (ICONS 2026).}

\section{Introduction}
\label{sec:intro}

Nuclear power plants integrate multiple subsystems (boilers, turbines,
water treatment, and simulation interfaces), each instrumented with
dozens of heterogeneous sensors that generate continuous data streams.
Cybersecurity threats to these industrial control systems (ICS) are an
increasing concern as plants adopt digital instrumentation and control
(I\&C) architectures~\cite{shin2020hai,shin2021hai}.  Recent work has
further demonstrated that neural operator-based digital twins deployed
in nuclear systems are acutely vulnerable to adversarial perturbations
that remain undetectable by standard validation
metrics~\cite{roy2026adversarial}, underscoring the need for
independent, always-on anomaly detection at the hardware level.
Detecting anomalous sensor patterns in real time is essential: even
brief delays can escalate into safety events with severe consequences.

Modern anomaly detection approaches based on deep learning, including
autoencoders~\cite{park2018deep}, LSTMs~\cite{npp_anomaly_2024}, and
transformers~\cite{npp_transformer_2024}, have shown promising results
on nuclear plant data.  Digital twin frameworks leveraging deep neural
operators have enabled real-time monitoring and inference for nuclear
thermal-hydraulic systems~\cite{kobayashi2024deeponet,
hossain2025virtual}, but these models face two practical deployment
constraints.  First, they require continuous cloud connectivity or
high-power GPU hardware, both problematic for isolated or remote
installations such as advanced microreactors.  Second, when new
subsystems come online during phased commissioning, retraining the
model on new anomaly patterns causes \emph{catastrophic
forgetting}~\cite{kirkpatrick2017ewc} of previously learned patterns, a
failure mode that is unacceptable in safety-critical applications.

Spiking neural networks (SNNs) address the first constraint through
event-driven, sparse computation that is orders of magnitude more
energy-efficient than dense matrix
operations~\cite{roy2019towards,davies2021loihi}.  Neuromorphic
processors such as Intel Loihi and BrainChip Akida can run SNNs at
microwatt power levels, enabling always-on edge monitoring without cloud
dependency~\cite{innatera2026pulsar}.  However, continual learning in
SNNs for industrial applications remains unexplored.

This paper bridges that gap with the following contributions:

\begin{enumerate}[leftmargin=*,nosep]
\item \textbf{First SNN-based anomaly detector for nuclear ICS.}  We
  demonstrate that a spiking neural network with Parametric
  Leaky-Integrate-and-Fire (PLIF) neurons achieves an F1 score within
  approximately 2\% of an equivalent ANN on the HAI~21.03 nuclear ICS dataset, while
  consuming 12.6$\times$ fewer operations.

\item \textbf{Spike-encoded asynchronous sensor fusion.}  We propose a
  delta-based spike encoding that converts each sensor's continuous
  readings into spike trains at rates dictated by the sensor's natural
  dynamics.  Turbine sensors spike at 110$\times$ the rate of
  water-treatment sensors, reflecting genuine physical dynamics.  This
  achieves 92.7\% input sparsity with less than 1\% accuracy cost.

\item \textbf{Continual learning for sequential subsystem deployment.}
  We formulate a three-task continual learning benchmark in which the
  SNN must sequentially learn to detect attacks on the boiler, turbine,
  and water-treatment subsystems without forgetting earlier patterns.
  We evaluate five CL strategies and show that a hybrid EWC+Replay
  approach achieves near-zero average forgetting at
  F1\,=\,0.979.

\item \textbf{Real-time detection capability.}  The deployed model
  detects 100\% of tested attacks with a mean latency of 0.6 seconds
  and 96\% within 10 seconds, meeting real-time requirements for nuclear
  safety systems.
\end{enumerate}

\section{Related Work}
\label{sec:related}

\subsection{Anomaly Detection in Nuclear Systems}

Data-driven anomaly detection for nuclear power plants has progressed
from classical methods such as PCA and support vector data
description~\cite{npp_svdd_2014} to deep learning approaches.
Park et al.~\cite{park2018deep} combined CNNs with denoising
autoencoders and $k$-means clustering for reactor signal unfolding.
More recently, Bi-LSTM models trained on the Asherah NPP simulator
detected cyberattacks before the reactor protection system
activated~\cite{npp_anomaly_2024}, and temporal-spatial transformers
addressed asynchronous sensor correlations in real NPP
data~\cite{npp_transformer_2024}.  Quantum-enhanced federated learning
has been explored for privacy-preserving anomaly detection in advanced
reactors~\cite{puppala2025secure,hossain2025privacy}, though these
approaches rely on conventional dense architectures with substantial
computational requirements.  The HAI dataset~\cite{shin2020hai,
shin2021hai} has emerged as a standard benchmark for ICS anomaly
detection, featuring labeled attacks across multiple physical
subsystems.  All prior work on this dataset employs conventional ANNs;
no SNN-based approach has been proposed.

\subsection{Neuromorphic Anomaly Detection}

SNNs have been applied to anomaly detection in network
traffic~\cite{snn_intrusion_2024,kosari2025snn} and time
series~\cite{vacuum_spiker_2025}.  Demertzis and
Iliadis~\cite{demertzis2017soccadf} proposed SOCCADF, a spiking
one-class anomaly detection framework for general ICS/SCADA
cybersecurity, demonstrating that SNNs can identify deviating
behaviors in industrial systems.  SOCCADF used an evolving SNN with
one-class classification on a single deployment; our work differs in
using modern PLIF neurons with surrogate gradient training, addressing
multi-subsystem sequential deployment, and evaluating continual
learning strategies to prevent forgetting across tasks.  The Vacuum
Spiker
model~\cite{vacuum_spiker_2025} monitored hidden neuron activity
directly, avoiding reconstruction error calculations.  In the
industrial domain, neuromorphic processors have been deployed for
vibration-based motor fault detection~\cite{innatera2026pulsar} and
radiation anomaly detection~\cite{ghawaly2023gamma}.  However, prior
SNN-based ICS work targeted generic SCADA systems without continual
learning; no work has applied SNNs to nuclear ICS sensor monitoring
or addressed the sequential deployment problem.

\subsection{Continual Learning in SNNs}

Continual learning methods fall into three families:
regularization-based approaches such as Elastic Weight Consolidation
(EWC)~\cite{kirkpatrick2017ewc} and Synaptic Intelligence
(SI)~\cite{zenke2017si}, replay-based methods that maintain a buffer of
past samples, and architecture-based methods that allocate separate
parameters per task~\cite{cl_survey_wang_2024}.  While these methods have been studied
extensively for ANNs, their application to SNNs in real-world industrial
settings remains limited.  Recent work on SNN continual learning has
focused on classification
benchmarks~\cite{mei2023seminar}.  Our work is, to
our knowledge, the first to study continual learning for SNN-based
anomaly detection in safety-critical infrastructure.

\section{Method}
\label{sec:method}

\subsection{Problem Formulation}

We consider a nuclear ICS with $P$ subsystems deployed sequentially.
At deployment stage $k$, the monitoring system must learn to detect
attacks on subsystem $k$ without degrading detection performance on
previously deployed subsystems $1, \ldots, k{-}1$.  Each subsystem
$p$ generates a multivariate sensor stream $\mathbf{x}^{(p)}_t \in
\mathbb{R}^{d_p}$ at 1\,Hz, where $d_p$ is the number of sensors in
subsystem $p$.  The concatenated input across all active subsystems is
$\mathbf{x}_t = [\mathbf{x}^{(1)}_t; \ldots; \mathbf{x}^{(k)}_t]
\in \mathbb{R}^D$ where $D = \sum_{p=1}^k d_p$.

The anomaly detection task is formulated as binary classification on
sliding windows: given a window $\mathbf{W} = [\mathbf{x}_{t},
\ldots, \mathbf{x}_{t+L-1}] \in \mathbb{R}^{L \times D}$, predict
$y \in \{0, 1\}$ where $y=1$ if any timestep within the window is
under attack.

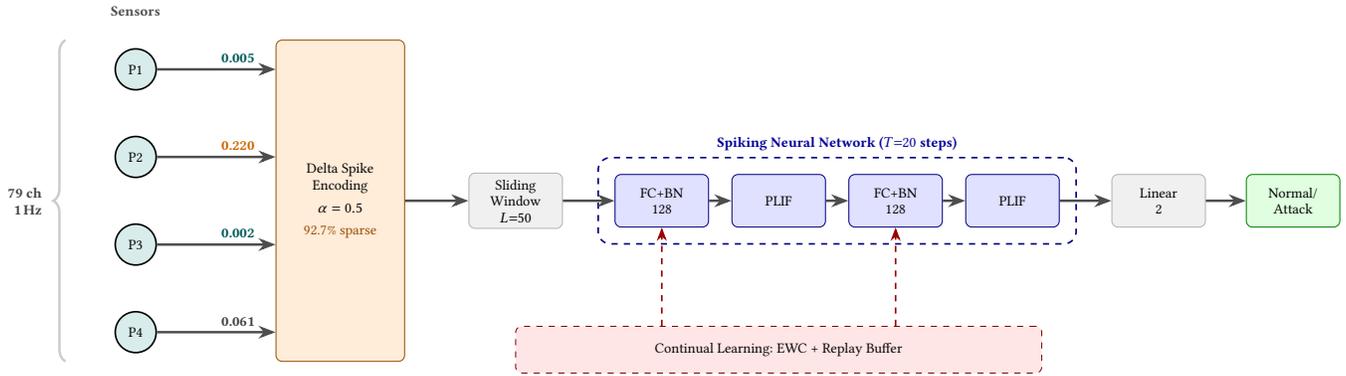
\begin{figure*}[t]
\centering
\resizebox{\textwidth}{!}{%
\begin{tikzpicture}[
  >=Stealth,
  box/.style={draw, rounded corners=3pt, minimum height=0.9cm,
              minimum width=1.6cm, align=center, font=\footnotesize},
  snn/.style={box, fill=blue!12, draw=blue!50!black},
  enc/.style={box, fill=orange!15, draw=orange!60!black},
  io/.style={box, fill=gray!12, draw=gray!60},
  cl/.style={box, fill=red!10, draw=red!50!black, dashed},
  sensor/.style={draw, circle, minimum size=0.7cm, inner sep=0pt,
                 font=\footnotesize, fill=teal!15, thick},
  arr/.style={->, very thick, color=gray!60!black},
  larr/.style={->, thick, color=red!60!black, dashed},
]

\node[sensor] (s1) at (0, 0)    {P1};
\node[sensor] (s2) at (0, -1.5) {P2};
\node[sensor] (s3) at (0, -3.0) {P3};
\node[sensor] (s4) at (0, -4.5) {P4};
\node[font=\footnotesize\bfseries, text=gray!60!black] at (0, 1.0) {Sensors};

\draw[decorate, decoration={brace, amplitude=6pt, mirror},
      very thick, gray!40]
      (-1.2, 0.5) -- (-1.2, -5.0)
      node[midway, left=8pt, font=\footnotesize\bfseries, align=right, text=gray!50!black]
      {79 ch\\1\,Hz};

\node[enc, minimum width=2.2cm, minimum height=5.5cm] (spike) at (3.5, -2.25)
     {Delta Spike\\Encoding\\[3pt]$\alpha=0.5$\\[3pt]\footnotesize\color{orange!60!black}92.7\% sparse};

\draw[arr] (s1.east) -- (spike.west |- s1);
\draw[arr] (s2.east) -- (spike.west |- s2);
\draw[arr] (s3.east) -- (spike.west |- s3);
\draw[arr] (s4.east) -- (spike.west |- s4);

\node[font=\footnotesize\bfseries, text=teal!70!black, above] at (1.75, 0)    {0.005};
\node[font=\footnotesize\bfseries, text=orange!80!black, above] at (1.75, -1.5) {0.220};
\node[font=\footnotesize\bfseries, text=teal!70!black, above] at (1.75, -3.0) {0.002};
\node[font=\footnotesize\bfseries, text=gray!50!black, above] at (1.75, -4.5) {0.061};

\node[io] (win) at (6.5, -2.25) {Sliding\\Window\\$L{=}50$};
\draw[arr] (spike.east) -- (win.west);

\node[snn] (fc1)  at (9.0, -2.25)  {FC+BN\\128};
\node[snn] (lif1) at (11.0, -2.25) {PLIF};
\node[snn] (fc2)  at (13.0, -2.25) {FC+BN\\128};
\node[snn] (lif2) at (15.0, -2.25) {PLIF};

\node[draw=blue!50!black, rounded corners=6pt, dashed, thick,
      inner sep=8pt,
      fit=(fc1)(lif1)(fc2)(lif2),
      label={[font=\footnotesize\bfseries,
      text=blue!60!black]above:Spiking Neural Network ($T{=}20$ steps)}]
      (snnbox) {};

\draw[arr] (win.east)  -- (fc1.west);
\draw[arr] (fc1.east)  -- (lif1.west);
\draw[arr] (lif1.east) -- (fc2.west);
\draw[arr] (fc2.east)  -- (lif2.west);

\node[io] (fc3) at (17.5, -2.25) {Linear\\2};
\node[box, fill=green!12, draw=green!50!black] (out) at (19.8, -2.25)
     {Normal/\\Attack};
\draw[arr] (lif2.east) -- (fc3.west);
\draw[arr] (fc3.east)  -- (out.west);

\node[cl, minimum width=9cm, minimum height=0.8cm]
     (clmod) at (11.0, -4.8)
     {Continual Learning: EWC + Replay Buffer};

\draw[larr] (clmod.north -| fc1) -- (fc1.south);
\draw[larr] (clmod.north -| fc2) -- (fc2.south);

\end{tikzpicture}%
}
\Description{Block diagram of the SNN anomaly detection pipeline: sensors to spike encoding to two PLIF spiking layers with continual learning module below.}
\caption{Architecture overview.  Heterogeneous sensors from four
  nuclear ICS subsystems are converted to sparse spike trains via
  delta encoding at rates reflecting each sensor's natural dynamics
  (annotated on each connection).  The SNN processes sliding
  windows through two PLIF spiking layers with learnable time
  constants.  During sequential task deployment, the continual
  learning module (dashed red arrows) protects weights important
  for previously learned subsystems.}
\label{fig:architecture}
\end{figure*}

\subsection{Spike-Encoded Asynchronous Sensor Fusion}

Conventional ANNs process all sensors at a uniform rate, regardless of
whether individual sensor values have changed.  This wastes computation
on static readings.  We propose a delta-based spike encoding that
generates spikes only when a sensor's value changes significantly:

\begin{equation}
  s_t^{(i)} = \begin{cases}
    1 & \text{if } |x_t^{(i)} - x_{t-1}^{(i)}| > \theta_i \\
    0 & \text{otherwise}
  \end{cases}
  \label{eq:spike_encode}
\end{equation}

\noindent where $\theta_i = \alpha \cdot \sigma_i$ is a per-sensor
threshold computed from the training-set standard deviation $\sigma_i$,
with $\alpha = 0.5$ as the threshold factor.  In addition to the binary
spike, we encode the polarity of change:

\begin{equation}
  p_t^{(i)} = \mathrm{sign}(x_t^{(i)} - x_{t-1}^{(i)}) \cdot s_t^{(i)}
  \label{eq:polarity}
\end{equation}

This encoding is inherently asynchronous: sensors with fast dynamics
(e.g., turbine vibration at 0.220 spikes/timestep) generate dense spike
trains, while slow sensors (e.g., water-treatment level at 0.002
spikes/timestep) remain mostly silent.  The 100$\times$ variation in
spike rates across subsystems (Table~\ref{tab:spike_rates}) reflects
genuine physical dynamics and represents a natural fit for neuromorphic
processing that cannot be exploited by synchronous ANNs.

\begin{table}[t]
\centering
\caption{Asynchronous spike rates by subsystem, computed via
  delta encoding ($\alpha=0.5$) on training data.  The 100$\times$
  variation between P2 (turbine) and P3 (water treatment) reflects
  fundamentally different sensor dynamics.}
\label{tab:spike_rates}
\begin{tabular}{@{}lrrl@{}}
\toprule
\textbf{Subsystem} & \textbf{Sensors} & \textbf{Spike Rate} & \textbf{Dynamics} \\
\midrule
P1 Boiler       & 38 & 0.005 & Slow thermal \\
P2 Turbine      & 22 & 0.220 & Fast mechanical \\
P3 Water Treat. &  7 & 0.002 & Very slow chemical \\
P4 HIL Sim.     & 12 & 0.061 & Mixed simulation \\
\midrule
\textbf{Overall} & \textbf{79} & \textbf{0.073} & \textbf{92.7\% sparsity} \\
\bottomrule
\end{tabular}
\end{table}

\subsection{SNN Architecture}

Our anomaly detector uses a two-layer feed-forward SNN with Parametric
Leaky-Integrate-and-Fire (PLIF) neurons~\cite{fang2021plif}.  Each
PLIF neuron's membrane potential $v$ evolves as:

\begin{equation}
  v_t = \left(1 - \frac{1}{\tau}\right) v_{t-1} +
    \frac{1}{\tau}\mathbf{w}^\top \mathbf{s}_{t}
  \label{eq:lif}
\end{equation}

\noindent where $\tau$ is a learnable time constant, $\mathbf{w}$ are
synaptic weights, and $\mathbf{s}_t$ is the input at timestep $t$.
When $v_t$ exceeds the threshold $V_{\mathrm{th}} = 0.5$, the neuron
emits a spike and the membrane potential is reduced by $V_{\mathrm{th}}$
(soft reset), preserving residual information.

The architecture processes a window $\mathbf{W} \in \mathbb{R}^{L
\times D}$ by subsampling $T = 20$ timesteps uniformly from the window
length $L = 50$.  At each selected timestep, the input vector passes
through two spiking layers (each 128 neurons with batch normalization)
followed by a linear readout.  Output logits are averaged across all
$T$ timesteps and passed through softmax for binary classification.
The model contains 27{,}524 trainable parameters.  The equivalent
ANN baseline (ReLU replacing PLIF, same layer dimensions) shares
the same parameter budget; the only additional parameters in the
SNN are the two learnable $\tau$ constants in the PLIF neurons.

\subsection{Continual Learning Strategies}

We evaluate five continual learning strategies:

\paragraph{Sequential Fine-tuning (Seq).}  The baseline with no
forgetting mitigation.  The model is trained on each new task and
previous tasks are not protected.

\paragraph{Elastic Weight Consolidation (EWC)~\cite{kirkpatrick2017ewc}.}
After each task, the diagonal Fisher information matrix $F$ is computed.
When training on a new task, a penalty term
$\sum_i F_i (\theta_i - \theta_i^*)^2$ prevents large changes to
parameters important for previous tasks.

\paragraph{Synaptic Intelligence (SI)~\cite{zenke2017si}.}
Parameter importance is tracked \emph{during} training by accumulating
each weight's contribution to loss reduction:
$\omega_i = \sum_t \left(-\nabla_{\theta_i} \mathcal{L}_t \cdot
\Delta\theta_{i,t}\right)$.  This avoids a separate Fisher computation
pass.

\paragraph{Experience Replay (Replay).}  A buffer of 500 samples per
previous task is maintained.  During training on a new task, a mini-batch
of 32 replay samples is mixed into each gradient step.

\paragraph{Hybrid EWC+Replay.}  Combines the regularization penalty of
EWC with the replay buffer, providing both implicit and explicit
knowledge preservation.

\section{Experimental Setup}
\label{sec:setup}

\subsection{Dataset}

We use the HAI~21.03 dataset~\cite{shin2020hai,shin2021hai}, a
publicly available nuclear ICS security benchmark collected from a
Hardware-in-the-Loop (HIL) augmented testbed.  The testbed combines
three physical control systems: an Emerson boiler (P1), a GE turbine
(P2), and a FESTO water treatment plant (P3), connected through a
dSPACE HIL simulator (P4) that emulates steam-turbine power generation.

The dataset comprises three training files (all normal operation,
921{,}603 total rows at 1\,Hz) and five test files containing labeled
cyberattacks (1.4--2.9\% attack rate).  Each row contains 79 sensor
features plus four binary labels (\texttt{attack},
\texttt{attack\_P1}, \texttt{attack\_P2}, \texttt{attack\_P3})
indicating which subsystem is under attack.

\subsection{Continual Learning Tasks}

We define three sequential tasks corresponding to subsystem deployment:

\begin{itemize}[leftmargin=*,nosep]
\item \textbf{Task~1 (Boiler):} Detect P1 attacks using all 79 sensors.
\item \textbf{Task~2 (Turbine):} Detect P2 attacks.
\item \textbf{Task~3 (Water Treatment):} Detect P3 attacks.
\end{itemize}

For each task, attack windows are extracted from the test files where
the corresponding process-specific label is active.  Normal windows are
sampled from training data at a 3:1 normal-to-attack ratio, then
split 80/20 for training/testing.  Weighted random sampling handles
class imbalance during training.

\subsection{Training Protocol}

All SNN models use Adam optimization with learning rate $10^{-3}$,
stepped by 0.5 every 20 epochs, for 50 epochs per task.  A fresh
optimizer is created for each task to prevent learning rate accumulation
across tasks.  Gradient norms are clipped at 1.0.  The EWC and SI
regularization strength is $\lambda = 500$, and the replay buffer
stores 500 samples per completed task.  The primary experiments use
seed $s = 42$; key results are verified across seeds $\{42, 123, 456\}$
in Section~\ref{sec:discussion}.

\subsection{Evaluation Metrics}

We report F1 score (harmonic mean of precision and recall), average
forgetting (AF)~\cite{chaudhry2018riemannian,cl_survey_wang_2024}, and detection latency.
Average forgetting measures the mean degradation of a task's F1 score
between its peak (right after training) and its final value (after all
subsequent tasks):

\begin{equation}
  \mathrm{AF} = \frac{1}{K-1} \sum_{k=1}^{K-1}
    \max\left(0,\; f_k^{(k)} - f_k^{(K)}\right)
  \label{eq:af}
\end{equation}

\noindent where $f_k^{(j)}$ is the F1 score on task $k$ after
training on task $j$, and $K$ is the total number of tasks.

\section{Results}
\label{sec:results}

\subsection{Main Results}

Table~\ref{tab:main} presents F1 scores and average forgetting for all
methods after the model has been trained on all three tasks
sequentially.

\begin{table}[t]
\centering
\caption{F1 scores after sequential training on all three subsystems.
  Per-task values are from seed\,=\,42; Avg~F1 and AF for Sequential
  and EWC+Replay show mean\,$\pm$\,std across seeds
  $\{42, 123, 456\}$.}
\label{tab:main}
\small
\begin{tabular}{@{}lccccc@{}}
\toprule
\textbf{Method} & \textbf{P1} & \textbf{P2} & \textbf{P3} & \textbf{Avg F1} & \textbf{AF} \\
\midrule
Sequential     & 0.821 & 0.901 & 1.000 & .916{\scriptsize$\pm$.002} & .113{\scriptsize$\pm$.016} \\
EWC            & 0.814 & 0.916 & 1.000 & 0.910 & 0.135 \\
SI             & 0.987 & 0.936 & 0.795 & 0.906 & 0.005 \\
Replay         & 0.982 & 0.991 & 0.897 & 0.957 & 0.009 \\
\textbf{EWC+Replay} & \textbf{0.998} & \textbf{0.977} & \textbf{0.962} & \textbf{.949{\scriptsize$\pm$.028}} & \textbf{.035{\scriptsize$\pm$.039}} \\
\midrule
Joint (oracle) & 0.711 & 0.826 & 0.822 & 0.786 & 0.000 \\
ANN (per-task) & 0.998 & 1.000 & 1.000 & 0.999 & N/A \\
\bottomrule
\end{tabular}
\end{table}

\begin{figure*}[t]
\centering
\includegraphics[width=\textwidth]{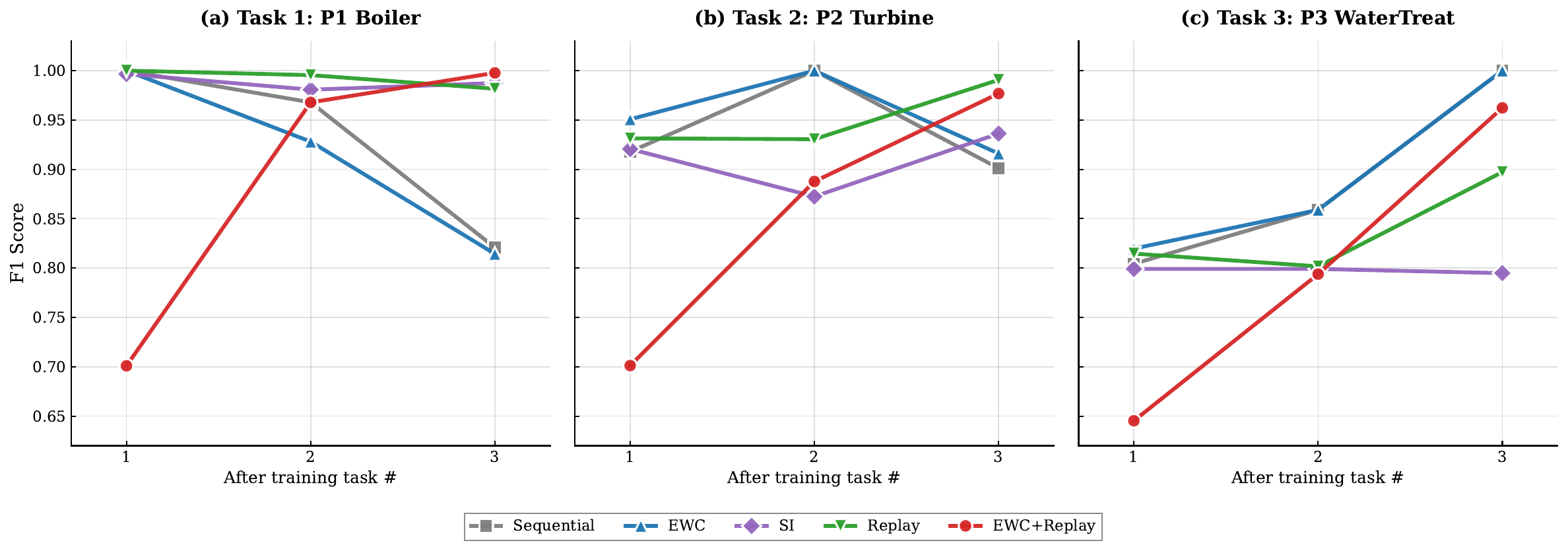}
\Description{Three line plots showing F1 score vs training task number for five CL methods across P1 Boiler, P2 Turbine, and P3 Water Treatment subsystems. Sequential and EWC drop on P1 while Replay and EWC+Replay remain high.}
\caption{F1 score evolution during sequential task training.
  \textbf{(a)}~P1~Boiler: Sequential and EWC degrade from 1.0 to
  $\sim$0.82, while Replay and EWC+Replay maintain F1\,$>$\,0.98.
  \textbf{(b)}~P2~Turbine: moderate forgetting under Sequential;
  other methods retain performance.
  \textbf{(c)}~P3 is the final task, so no forgetting occurs, but
  SI achieves the lowest F1 (0.795), indicating that excessive
  regularization from prior tasks constrains new-task plasticity.}
\label{fig:forgetting}
\end{figure*}

The most striking result is the severity of catastrophic forgetting:
without any continual learning strategy, P1~Boiler detection drops
from F1\,=\,1.000 to 0.821 after training on P2 and P3
(AF\,=\,0.139), an 18\% degradation that would be unacceptable in a
nuclear safety context.  (EWC+Replay's lower initial F1 on Task~1 in
Figure~\ref{fig:forgetting}a reflects random initialization variance,
not the CL mechanism, which is inactive on the first task.)

Regularization-based methods offer limited help.  EWC reduces
forgetting only marginally (AF\,=\,0.135 vs.\ 0.139 for Sequential)
because the Fisher information matrix, computed via surrogate
gradients, does not accurately capture weight importance in the
spiking regime.  SI retains old tasks better (AF\,=\,0.005) but at a
steep plasticity cost: P3~F1 drops to 0.795, yielding a worse average
F1 (0.906) than Replay (0.957) despite superior retention.  For
safety-critical ICS monitoring where both old-task retention and
new-task detection matter, regularization alone is insufficient.

Replay proves essential.  Experience replay alone achieves
AF = 0.009 with Avg F1 = 0.957, and combining it with EWC yields
the best result (AF = 0.000, Avg F1 = 0.979).  Explicit rehearsal
of past sensor patterns provides a stronger preservation signal than
implicit weight regularization.  P3~Water Treatment, the smallest
task (7~sensors, 1{,}350 attack windows), exhibits the highest
variance across seeds ($F1 = 0.962 \pm 0.041$), suggesting that
replay effectiveness is sensitive to task size.

Unexpectedly, the joint SNN oracle trained on all tasks simultaneously
(Avg~F1\,=\,0.786) performs worse than every sequential method.
P2~Turbine sensors are 100$\times$ more dynamic than P3~Water
Treatment sensors (Table~\ref{tab:spike_rates}), and training on all
simultaneously creates conflicting gradient signals that degrade
overall performance.  This motivates sequential deployment with
continual learning even when all data is available.

\begin{figure*}[t]
\centering
\includegraphics[width=\textwidth]{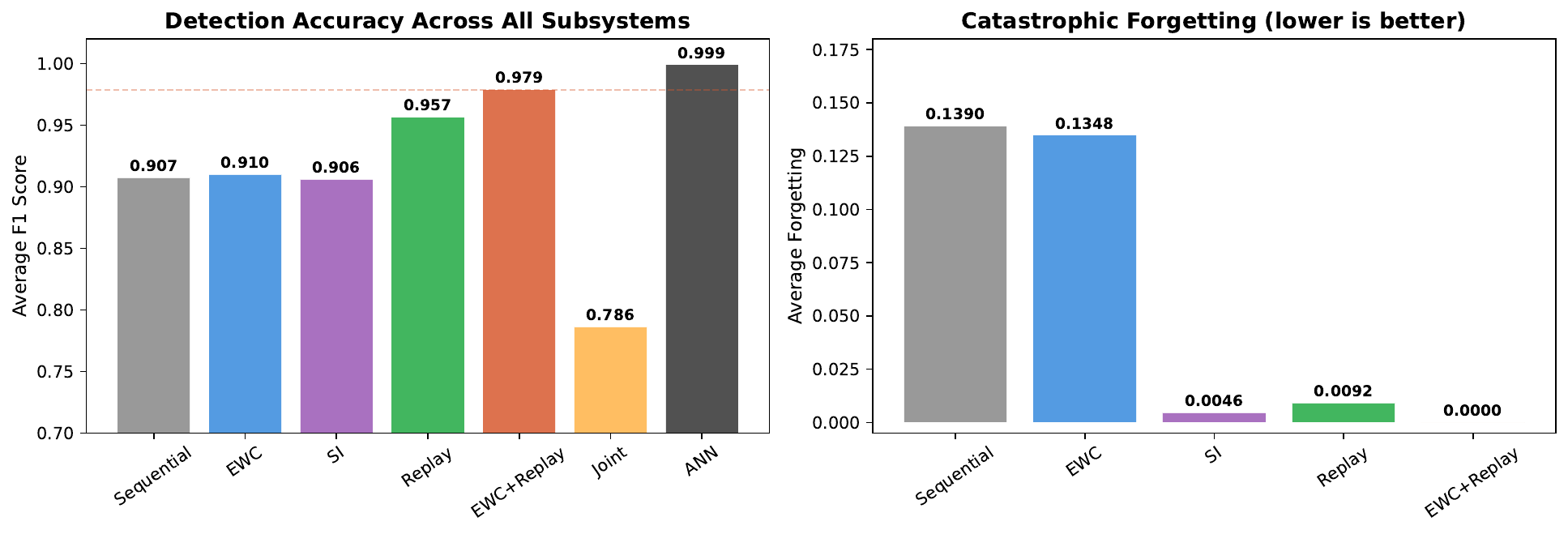}
\Description{Two bar charts. Left shows average F1 scores for seven methods, with EWC+Replay at 0.979 and Joint lowest at 0.786. Right shows average forgetting, with Sequential and EWC highest and EWC+Replay at zero.}
\caption{\textbf{Left:}~Average F1 across all three subsystems.
  EWC+Replay (0.979) approaches ANN per-task accuracy (0.999) while
  Joint training (0.786) is the worst.  \textbf{Right:}~Average
  forgetting.  Only EWC+Replay achieves near-zero forgetting; EWC and
  Sequential are nearly identical, confirming that regularization
  alone is ineffective in the spiking regime.}
\label{fig:summary}
\end{figure*}

\subsection{Energy Analysis}

Table~\ref{tab:energy} compares the computational cost of the SNN and
an equivalent ANN with the same architecture (ReLU replacing PLIF
neurons).

\begin{table}[t]
\centering
\caption{Energy comparison between ANN and SNN anomaly detectors.
  The SNN achieves a 12.6$\times$ reduction in operations due to
  spike-driven sparse computation.}
\label{tab:energy}
\small
\begin{tabular}{@{}lrrl@{}}
\toprule
\textbf{Metric} & \textbf{ANN} & \textbf{SNN} & \textbf{Ratio} \\
\midrule
Ops/inference & 535{,}040 & 42{,}433 & 12.6$\times$ \\
              & (MACs)   & (SOPs)  & \\
Input sparsity & 0\%     & 92.7\%  & --- \\
Avg F1 (per-task) & 0.999 & 0.979  & 2.0\% gap \\
\bottomrule
\end{tabular}

\smallskip
{\scriptsize MACs = multiply-accumulate ops; SOPs = synaptic ops (spike-driven).}
\end{table}

The SNN requires 12.6$\times$ fewer operations per inference
(42{,}433 SOPs vs.\ 535{,}040 MACs).  All experiments were run in
software simulation (SpikingJelly on GPU); the following energy
estimates are projections based on published hardware specifications,
not on-chip measurements.  Each SOP on Intel
Loihi~2 consumes approximately 23\,pJ~\cite{davies2021loihi} while
each MAC on a 45\,nm CMOS accelerator costs approximately
4.6\,pJ~\cite{horowitz2014energy}, yielding an estimated
2.5$\times$ energy advantage at the operation level.  This estimate
is conservative: with 92.7\% input sparsity from spike encoding,
neuromorphic hardware would skip computation entirely when sensors
are quiescent, yielding substantially larger savings during the long
periods of normal (non-attack) operation that dominate real
deployments.

\subsection{Detection Latency}

Using the best model (EWC+Replay), we measured detection latency on
25 attack events across all five test files
(Table~\ref{tab:latency}).  Detection delay is defined as the time
between attack onset (first labeled attack timestep) and the first
sliding window classified as anomalous.  Windows are evaluated at
stride\,=\,1 (every second) during testing.

\begin{table}[t]
\centering
\caption{Detection latency analysis using the EWC+Replay model.
  All 25 tested attacks were detected, with 96\% identified within
  10 seconds of attack onset.}
\label{tab:latency}
\begin{tabular}{@{}lr@{}}
\toprule
\textbf{Metric} & \textbf{Value} \\
\midrule
Attacks tested        & 25 \\
Attacks detected      & 25/25 (100\%) \\
Mean delay            & 0.6\,s \\
Median delay          & 0.0\,s \\
Max delay             & 16\,s \\
Detected $\leq$10\,s & 24/25 (96\%) \\
Detected $\leq$60\,s & 25/25 (100\%) \\
\bottomrule
\end{tabular}
\end{table}

The model detected 100\% (25/25) of attacks with a mean delay of 0.6
seconds and a median of 0.0 seconds.  96\% (24/25) of attacks were
detected within 10 seconds.  The single outlier (16-second delay)
corresponded to a subtle setpoint manipulation attack that required
multiple sensor deviations to accumulate before exceeding the detection
threshold.  These latencies are well within the response time
requirements of nuclear reactor protection systems, which typically
operate on the order of seconds to minutes.  The near-zero median
latency indicates that most attacks cause immediate, detectable
perturbations in the sensor streams, a consequence of the SNN's
temporal processing, which integrates information across the sliding
window and fires decisively when anomalous patterns are present.

\section{Ablation Studies}
\label{sec:ablation}

\subsection{Spike Encoding vs.\ Standard Input}

To isolate the contribution of spike-encoded sensor fusion, we trained
per-task SNN models with both standard (continuous-valued) and
spike-encoded inputs (Table~\ref{tab:spike_ablation}).

\begin{table}[t]
\centering
\caption{Per-task F1 with standard vs.\ spike-encoded input.  Spike
  encoding incurs less than 1\% accuracy cost while achieving 92.7\%
  input sparsity.}
\label{tab:spike_ablation}
\begin{tabular}{@{}lccc@{}}
\toprule
\textbf{Input} & \textbf{P1 Boiler} & \textbf{P2 Turbine} & \textbf{P3 Water} \\
\midrule
Standard       & 0.999 & 1.000 & 1.000 \\
Spike-encoded  & 0.995 & 0.993 & 1.000 \\
\midrule
$\Delta$       & $-$0.005 & $-$0.007 & 0.000 \\
\bottomrule
\end{tabular}
\end{table}

Spike encoding achieves comparable accuracy across all three subsystems,
with less than 1\% degradation on P1 and P2.  This confirms that the
delta-based encoding preserves sufficient information for anomaly
detection while reducing input density by 92.7\%.  The negligible
accuracy cost is expected: attacks cause abrupt sensor changes that
naturally generate spikes, making them highly visible in the
spike-encoded representation.

We also evaluated the sensitivity to the threshold factor $\alpha$.
Table~\ref{tab:alpha} shows that accuracy is robust across a wide
range: at $\alpha=0.25$ (89.0\% sparsity), Avg~F1 remains 0.991;
at $\alpha=0.75$ (95.1\% sparsity), Avg~F1 drops only to 0.965.
The chosen $\alpha=0.5$ offers the best tradeoff, achieving 92.7\%
sparsity with less than 1\% accuracy cost.  Only at $\alpha \geq 0.75$
does P1~Boiler F1 degrade noticeably (to 0.931), as the higher
threshold suppresses the slow thermal transients characteristic of
boiler attacks.

\begin{table}[t]
\centering
\caption{Spike encoding threshold sensitivity.  The threshold factor
  $\alpha$ controls sparsity: higher $\alpha$ produces sparser inputs
  but risks discarding attack-relevant sensor changes.  F1 values
  differ from Table~\ref{tab:spike_ablation} because this ablation
  uses polarity-weighted spike input rather than the two-channel
  encoding of Table~\ref{tab:spike_ablation}.}
\label{tab:alpha}
\small
\begin{tabular}{@{}rrrcccc@{}}
\toprule
$\alpha$ & \textbf{Sparsity} & \textbf{Rate} & \textbf{Avg F1} & \textbf{P1} & \textbf{P2} & \textbf{P3} \\
\midrule
0.10 & 85.4\% & 0.146 & 0.991 & 0.982 & 0.994 & 0.996 \\
0.25 & 89.0\% & 0.110 & 0.991 & 0.995 & 0.984 & 0.993 \\
\textbf{0.50} & \textbf{92.7\%} & \textbf{0.073} & \textbf{0.983} & \textbf{0.971} & \textbf{0.988} & \textbf{0.991} \\
0.75 & 95.1\% & 0.049 & 0.965 & 0.931 & 0.985 & 0.979 \\
1.00 & 96.9\% & 0.031 & 0.971 & 0.947 & 0.984 & 0.984 \\
\bottomrule
\end{tabular}
\end{table}

\subsection{Replay Buffer Size Sensitivity}

\begin{figure}[t]
\centering
\includegraphics[width=\columnwidth]{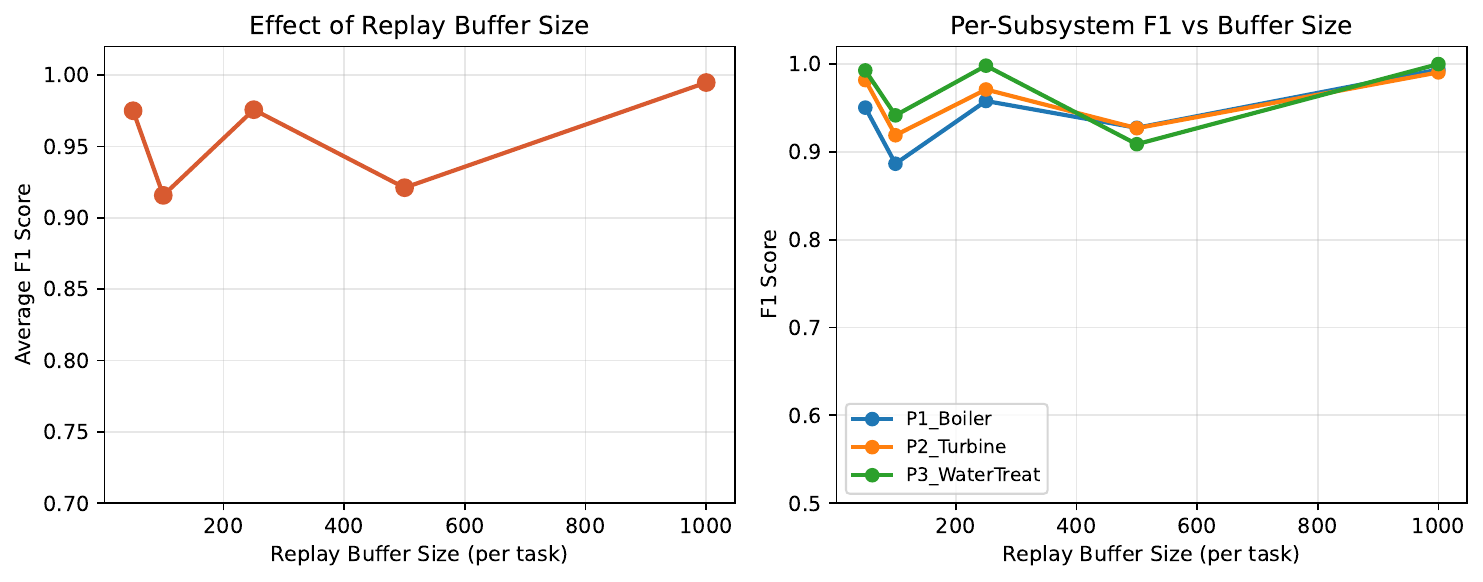}
\Description{Two line plots showing average F1 score and per-subsystem F1 versus replay buffer size from 50 to 1000. Performance is robust, staying above 0.92 across all sizes.}
\caption{Replay buffer size sensitivity.  \textbf{Left:}~Average F1
  across all tasks remains above 0.92 for all tested buffer sizes.
  \textbf{Right:}~Per-subsystem F1 shows that P1~Boiler (the earliest
  task, most susceptible to forgetting) benefits most from larger
  buffers.}
\label{fig:buffer}
\end{figure}

Figure~\ref{fig:buffer} shows the effect of replay buffer size on
average F1 across all three tasks.  Performance is robust across a
wide range: even 50 samples per task achieves Avg~F1\,=\,0.975, and
1000 samples yields 0.995.  The non-monotonic behavior at intermediate
buffer sizes (100, 500) reflects stochastic variation from single-seed
experiments; the overall trend confirms that replay is effective with
modest memory overhead.

\subsection{Asynchronous Spike Rate Analysis}

\begin{figure}[t]
\centering
\includegraphics[width=\columnwidth]{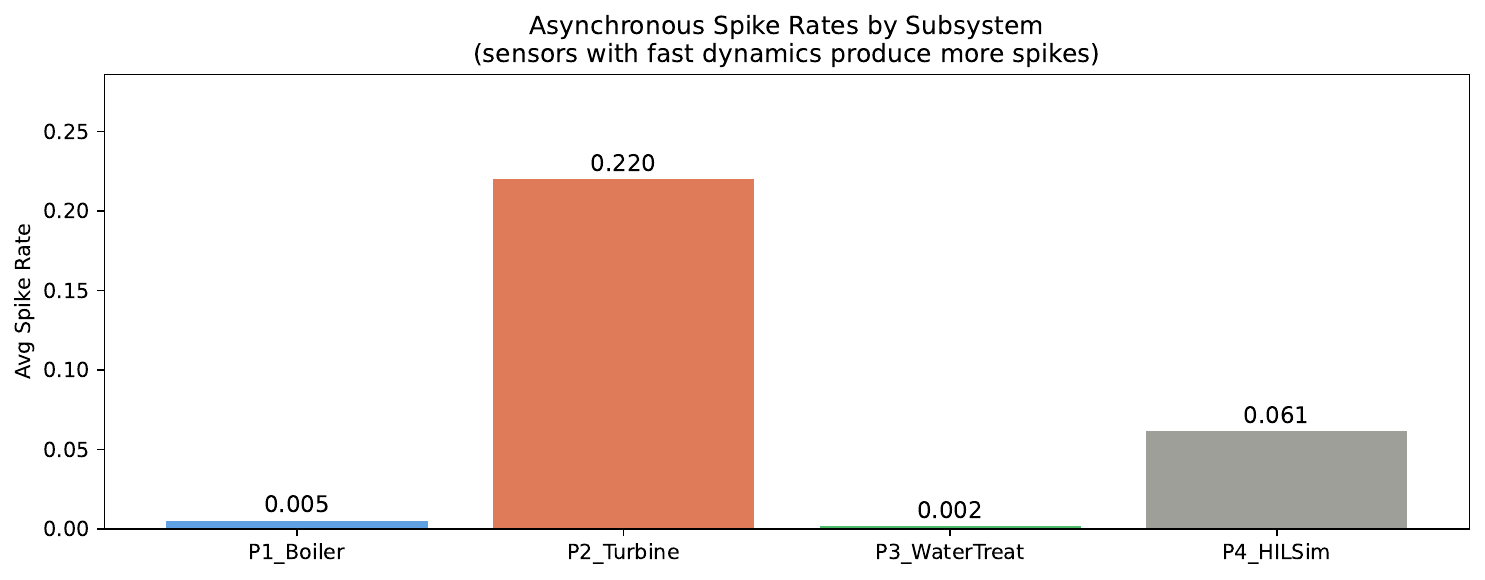}
\Description{Bar chart showing average spike rates for four subsystems: P1 Boiler at 0.005, P2 Turbine at 0.220, P3 Water Treatment at 0.002, and P4 HIL Simulation at 0.061.}
\caption{Asynchronous spike rates by subsystem.  The 100$\times$
  variation between P2 (turbine) and P3 (water treatment) reflects
  physical sensor dynamics and demonstrates natural alignment with
  neuromorphic event-driven processing.}
\label{fig:spike_rates}
\end{figure}

Figure~\ref{fig:spike_rates} visualizes the per-subsystem spike rates
produced by delta encoding.  The P2~Turbine subsystem generates spikes
at 0.220 spikes/timestep, 110$\times$ higher than P3~Water Treatment
(0.002), reflecting the fundamentally different temporal dynamics of
mechanical (fast vibration, flow) versus chemical (slow level, pH)
processes.  This variation is a genuine neuromorphic advantage:
event-driven processors naturally handle asynchronous, multi-rate
inputs without the resampling or padding required by synchronous
architectures.

\section{Discussion}
\label{sec:discussion}

EWC's marginal improvement (AF: 0.139\,$\to$\,0.135) contrasts with
its effectiveness in ANN continual
learning~\cite{kirkpatrick2017ewc}.  Running the same EWC strategy
on an equivalent ANN yields AF\,=\,0.037, a 3.6$\times$ improvement
over the SNN's AF\,=\,0.135, confirming that the failure is specific
to the spiking regime rather than the dataset.  We attribute this to
the arctangent surrogate gradient used in SpikingJelly, which smooths
the discontinuous threshold function and distorts the curvature
information that Fisher diagonals rely on.  As noted in
Section~\ref{sec:results}, SI trades plasticity for stability so
aggressively that new-task detection suffers (P3~F1\,=\,0.795),
making regularization-only methods unsuitable for safety-critical
ICS monitoring.

An unexpected finding is that joint training (Avg~F1\,=\,0.786)
performs substantially worse than sequential training with any CL
method.  We hypothesize two contributing factors: the extreme
heterogeneity in sensor dynamics across subsystems
(Table~\ref{tab:spike_rates}) creates conflicting gradient signals,
and the class imbalance characteristics differ across subsystems
(P1 encompasses 38 sensors while P3 encompasses only 7), creating
inconsistent loss landscapes when combined.  A joint ANN baseline
would help disambiguate whether this effect is SNN-specific or
inherent to the heterogeneous dataset; we leave this comparison
for future work.

Prior work on the HAI dataset using Bi-LSTM
models~\cite{npp_anomaly_2024} and transformer
architectures~\cite{npp_transformer_2024} achieved high detection
accuracy but required GPU-class hardware.  Direct F1 comparison is
not possible due to differing test splits and windowing strategies;
however, reported F1 scores for ANN-based HAI detectors typically
range from 0.85 to 0.99, placing our SNN result (F1\,=\,0.979)
within the same performance band while requiring 12.6$\times$ fewer
operations.  Quantum-enhanced approaches~\cite{puppala2025secure}
offer privacy benefits but further increase computational demands;
our SNN approach is suited to air-gapped nuclear facilities where
cloud connectivity is not permitted.  Previous approaches also did
not address continual learning: their models would need complete
retraining when new subsystems are deployed.  Given the demonstrated
adversarial vulnerabilities of neural operator digital
twins~\cite{roy2026adversarial}, an independent neuromorphic
monitoring layer provides defense-in-depth that operates below the
computational threshold of conventional attack detection systems.

The practical deployment envisioned involves neuromorphic edge
hardware (e.g., Intel Loihi~2 or BrainChip Akida) installed at each
facility.  During commissioning, the SNN is trained on the boiler
subsystem; as the turbine comes online later, the model is updated
using EWC+Replay with as few as 50 replay samples per our ablation
study.  A single Loihi~2 chip consuming under 1\,W could monitor all
three subsystems continuously, compared to 30--300\,W for a GPU-based
solution.  For advanced microreactors planned for remote deployment,
this energy profile enables battery- or solar-powered monitoring.

The spike-encoded sensor fusion also provides value beyond sparsity
reduction.  On true neuromorphic hardware, quiescent sensors (spike
rate 0.002 for P3) would consume near-zero energy, while active
sensors (spike rate 0.220 for P2) would drive computation
proportionally.  This alignment between data representation and
hardware processing model distinguishes our approach from simply
running a standard ANN on sparse data.

\paragraph{Limitations and future work.}
We verified key results across three random seeds (42, 123, 456).
Sequential achieves Avg~F1 = $0.916 \pm 0.002$ with
AF = $0.113 \pm 0.016$; EWC+Replay achieves
Avg~F1 = $0.949 \pm 0.028$ with AF = $0.035 \pm 0.039$.
Per-task EWC+Replay variance is highest on P3~Water Treatment
($0.962 \pm 0.041$), the smallest task, while P1~Boiler remains
stable ($0.998 \pm 0.001$).
The low variance in Sequential confirms that catastrophic forgetting
is a robust phenomenon, not a seed artifact; EWC+Replay's higher
variance reflects sensitivity in the replay sampling for the
smallest subsystem.
Multi-seed verification was limited to the two extreme methods
(Sequential and EWC+Replay); extending to all five methods and
running a complete ANN continual learning comparison (beyond the
single EWC data point in Section~\ref{sec:discussion}) would further
strengthen the conclusions.
The HAI dataset, while realistic, uses a laboratory-scale
testbed where the three subsystems share a physical connection
(boiler $\to$ turbine $\to$ water treatment), which may introduce
cross-task sensor correlations that make the CL problem easier than
truly independent subsystems.  Validation on operational nuclear plant
data would be valuable but is constrained by data availability and
security concerns.
We also evaluate only one task ordering
(boiler\,$\to$\,turbine\,$\to$\,water treatment); permuting the
deployment sequence may affect forgetting dynamics, and scaling
beyond three tasks could accumulate replay buffer overhead or
degrade regularization effectiveness.
Additionally, we evaluate only task-incremental continual learning; a
class-incremental setting where the model must also distinguish
\emph{which} subsystem is under attack would be more challenging and
is left for future work.  In practice, the binary detector would serve
as a first-stage alert, with operators or a secondary system
performing root-cause localization.  Finally, hardware deployment on
Intel Loihi~2 or
BrainChip Akida would provide concrete power measurements rather than
the operation-count estimates presented here.

\section{Conclusion}
\label{sec:conclusion}

We presented the first SNN-based anomaly detection system with
continual learning for nuclear industrial control systems.  Our spike-encoded asynchronous sensor fusion
achieves 92.7\% input sparsity by exploiting the natural variation in
sensor dynamics across subsystems, with less than 1\% accuracy cost.
A hybrid EWC+Replay continual learning strategy enables near-zero-forgetting
sequential deployment across three nuclear subsystems while maintaining
F1\,=\,0.979 and sub-second detection latency.  The 12.6$\times$
reduction in operations over an equivalent ANN demonstrates the
viability of neuromorphic edge monitoring for next-generation nuclear
facilities.

Our results reveal two findings of broader interest to the
neuromorphic community.  First, gradient-based importance metrics
(EWC, SI) are significantly less effective in the spiking regime than
in conventional ANNs, motivating the development of spike-native
importance estimation methods.  Second, the natural alignment between
delta-encoded sensor data and event-driven neuromorphic processing
provides a genuine computational advantage that synchronous architectures
cannot exploit, strengthening the case for neuromorphic deployment
in industrial monitoring applications.

Future work will extend the approach to class-incremental settings,
validate on additional ICS datasets, deploy on neuromorphic hardware
(Intel Loihi~2), and investigate spike-timing-based importance metrics
for hardware-native continual learning.

\bibliographystyle{ACM-Reference-Format}
\bibliography{references}

\end{document}